\documentclass[10pt,twocolumn,letterpaper]{article}

\usepackage{iccv}
\usepackage{times}
\usepackage{epsfig}
\usepackage{graphicx}
\usepackage{amsmath}
\usepackage{amssymb}
\usepackage{multirow}
\usepackage[accsupp]{axessibility}

\usepackage[pagebackref=true,breaklinks=true,bookmarks=false,colorlinks=true]{hyperref}

\iccvfinalcopy 

\def\iccvPaperID{****} 

\ificcvfinal\fi

\begin{document}

\title{Few shot font generation via transferring similarity guided global style and quantization local style}

\author{
Wei Pan$^{1}$\quad
Anna Zhu$^{1}$\thanks{Corresponding author}\quad
Xinyu Zhou$^{1}$\quad
Brian Kenji Iwana$^{2}$\quad
Shilin Li$^{1}$\quad
\\
$^{1}$ School of Computer Science and Artificial Intelligence, Wuhan University of Technology
\\
$^{2}$ Human Interface Laboratory, Kyushu University
\\
{\tt\small \{aaawei, annazhu, 297932, shilinli\}@whut.edu.cn} \quad
{\tt\small iwana@ait.kyushu-u.ac.jp}
}

\maketitle
\ificcvfinal\thispagestyle{empty}\fi

\begin{abstract}
  Automatic few-shot font generation (AFFG), aiming at generating new fonts with only a few glyph references, reduces the labor cost of manually designing fonts. However, the traditional AFFG paradigm of style-content disentanglement cannot capture the diverse local details of different fonts. So, many component-based approaches are proposed to tackle this problem. The issue with component-based approaches is that they usually require special pre-defined glyph components, e.g., strokes and radicals, which is infeasible for AFFG of different languages. In this paper, we present a novel font generation approach by aggregating styles from character similarity-guided global features and stylized component-level representations. We calculate the similarity scores of the target character and the referenced samples by measuring the distance along the corresponding channels from the content features, and assigning them as the weights for aggregating the global style features. To better capture the local styles, a cross-attention-based style transfer module is adopted to transfer the styles of reference glyphs to the components, where the components are self-learned discrete latent codes through vector quantization without manual definition. With these designs, our AFFG method could obtain a complete set of component-level style representations, and also control the global glyph characteristics. The experimental results reflect the effectiveness and generalization of the proposed method on different linguistic scripts, and also show its superiority when compared with other state-of-the-art methods. The source code can be found at \href{https://github.com/awei669/VQ-Font}{https://github.com/awei669/VQ-Font}. 
\end{abstract}

\section{Introduction}

Font design techniques can benefit many critical applications, such as logo designs, data augmentation for text-related tasks, handwriting imitation and identification, etc. However, traditional font design heavily depends on expert designers rendering the glyph styles for each character manually, making the creation of fonts extremely expensive and labor-intensive, especially for glyph-rich scripts.

Recently, with the development of deep learning techniques, many automatic few-shot font generation (AFFG) methods have been proposed. They have been created using Convolution Neural Networks (CNNs)~\cite{CNN}, Generative Adversarial Networks (GANs)~\cite{GAN}, Transformers~\cite{T2TViT}, etc. The AFFG methods use only a few reference font images for generating different glyphs automatically. The typical strategy follows the style and content disentanglement and combination paradigm~\cite{8784736,Srivatsan2019ADF}, and either adopts global style representation or component-wise style representation. The global style representation~\cite{Jiang2019SCFontSC,8784736} is learned and extracted from all the references of each style, which could capture the global characteristics, such as character size and stroke space. However, it lacks the representation of diverse local details, such as the shape and length of local strokes and serif size. On the contrary, the component-wise style representation category~\cite{RD-GAN,cha2020few,park2021multiple} generally decomposes each reference sample sharing the same font into pre-defined components and radicals. It either conditions the style encoders jointly on the glyph image and the corresponding component labels or adopts component-label classification losses to train the style encoder. This can be infeasible because each character should be manually associated with a certain set of components, which requires more preparation when applying for new scripts. Additionally, for different content images, their local relations with reference samples can vary. That means the local style representations for different content characters are required to be recomputed when given fixed reference samples, which increases the computational cost.

To tackle the above issue, we propose a hybrid global and local style transferring approach for AFFG in this paper. Since the global style representation of fonts controls more intra-style consistent properties, e.g., the locations, sizes, stroke thickness, and spaces of characters, while the local style representation focuses on capturing inter-style inconsistent component details, e.g., stroke shape, serif-ness, stroke deformation. Therefore, we leverage both the global and local styles for feature complementation. In order to obtain the global style feature representation, we calculate the similarity scores of the target glyph and the referenced samples by measuring their content feature distances, and then assign them as the weights for aggregating the style features. For local style feature representation, the glyph components are first learned automatically, which are discrete latent codes decomposed from a set of glyphs by vector quantization. Then, a cross-attention transformer is employed to transfer component-wise styles, with the representation of the learned components as the queries and the style representations of the reference glyphs as the keys and values. Contrastive learning is used to learn the local styles in an unsupervised way. For each forward pass, the styles from the reference samples can be transferred onto all the components. So, this local style extraction process is independent of the content glyph, avoiding multiple component-wise representation calculations for different inputs. Finally, the global and local style representations are combined with content features, and then decoded into the target glyph. Moreover, we adopt GAN and a self-reconstruction strategy for training the model without strong supervision. Therefore, it can be easily applied for different script font generation.

We demonstrate the effectiveness of the proposed method on the Chinese mainly. The experimental results reflect the necessity of combing global and local representations, and also tell that our method outperforms other state-of-the-art (SOTA) AFFG methods given very limited reference examples.

In summary, the contributions of this paper are as follows:
\begin{itemize}
    \item We propose a novel AFFG method leveraging complementary global and local representations, which is able to capture intra-style consistent properties and intra-style inconsistent structures of reference glyphs. 
    \item Similarity of content is used to obtain global styles. It takes a similar degree of glyph structures into consideration. This strategy can better transfer styles for glyphs owning the same components with reference.
    \item Pre-trained Vector Quantization-based Variational Autoencoder (VQ-VAE) is adopted to extract components automatically, component labels are not required. The local styles can be transferred to all the components via cross-attention in one-forward pass. it is efficient for font library creation because it is content irrelevant. A style contrastive loss is proposed to unsupervised transfer the component-level styles.
    \item Experimental results show great generalizability of our model for unseen fonts, unseen characters, and different scripts. It achieves SOTA performance for font generation even with very limited reference samples. Additionally, it can transfer styles onto cross-linguistic in the zero-shot manner.
\end{itemize}


\section{Related Works}

\textbf{Image-to-Image Translation.}
Image-to-image (I2I) translation aims to learn a mapping between source and target domains while retaining the content of the source. GAN-based methods~\cite{isola2017image, choi2018stargan, liu2018unified} were widely used in this field. Pix2pix~\cite{isola2017image} first adopted conditional GAN into the I2I task with paired data. CycleGAN~\cite{zhu2017unpaired} introduced circular consistency loss for unsupervised I2I translation~\cite{bousmalis2017unsupervised,zhu2017unpaired}. Recently, several I2I methods~\cite{baek2021rethinking, bhattacharjee2020dunit} were proposed to tackle multi-class unsupervised I2I translation problems, aiming to simultaneously generate multiple style outputs given the same input.

Font generation belongs to the typical I2I translation task, so the generic I2I translation approaches could be adaptively modified for font generation~\cite{FUNIT,Gao2020GANBasedUC}.

\textbf{Character Style Transfer.}
Character style transfer mainly focused on typography transfer. Traditional methods constructed the transformation through shape modeling~\cite{Xu2009AutomaticGO} and statistical modeling~\cite{4804823}. Recent approaches applied deep learning techniques for character synthesis~\cite{Baluja2017LearningTS}. Some methods were able to synthesize glyphs as well as the textural effects~\cite{MC-GAN,9098082,AGIS}. However, the font generation task emphasizes more on the consistency of character shape instead of text effects.

\textbf{Global Style-content Disentanglement for Font Generation.}
Global style-content disentanglement and recombination is a popular strategy for font generation. It models each font style as a universal representation. EMD~\cite{zhang2018separating} proposed a network architecture with two encoders, one for content and the other for style. They generated arbitrary fonts by mixing the content and style features, followed by decoding the mixed features. MDM \cite{zhu2022text} built a framework to disentangle the text images into three factors: text content, font, and style features, and then remixed the factors of different images to transfer a new style. ELDF \cite{lee2022arbitrary} utilized new consistency losses that forced any combination of encoded features of the stacked inputs to own the consistent features of text contents and font styles.

However, the local details for the font are more diverse and are hard to capture only by global style representation.

\begin{figure*}
    \centering
    \includegraphics[width=\linewidth]{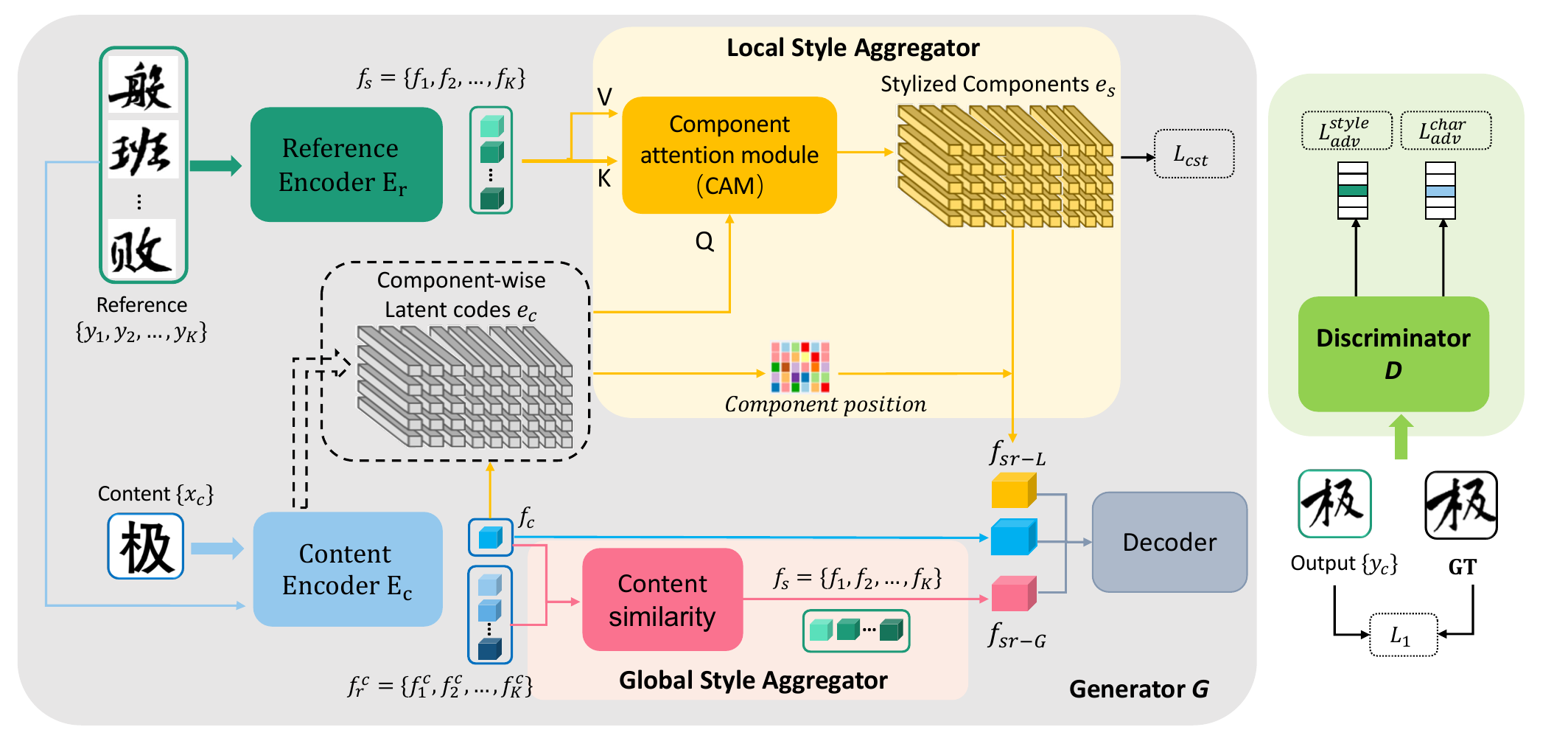}
    \caption{The architecture of our model. The generator consists of five parts: a pre-trained content encoder, a reference style encoder (marked in dark green), a local style aggregator via CAM (marked in yellow), a global style aggregator with content similarity guidance (marked in pink), and a decoder combining content and style features and style features for font generation (marked in gray). A discriminator (marking green) is followed to distinguish the real and fake images, and it simultaneously classifies the content and style category of the generated character.}
    \label{fig_1}
\end{figure*}

\textbf{Component-based Font Generation.}
Since characters such as Chinese characters can be decomposed, some component-based font generation methods~\cite{cha2020few, park2021few, park2021multiple} have emerged recently. They do not encode the entire image to extract style information, but instead extract style information for each component that makes up a syllable and use it to generate fonts. For instance, CalliGAN \cite{wu2020calligan} decomposed characters into components and offered low-level structure information, including the order of strokes, to guide the generation process. SA-VAE \cite{sun2018learning} disentangled the style and content as two irrelevant domains with encoding Chinese characters into high-frequency character structure configurations and radicals. Tang et al.~\cite{Tang2022FewShotFG} proposed FS-Font model that adopted a cross-attention mechanism to aggregate the reference styles into a fine-grained style representation. XMP-Font~\cite{Liu2022XMPFontSC} proposed a self-supervised cross-modality pre-training cross-modality transformer network to model style representations of stroke-level, component-level, and character-level. 

The component-based font generation methods significantly improve the generated glyphs’ quality with few-shot references. However, most of them require you to pre-define the radicals or components manually.

\section{Method}
In this section, we introduce the overall framework, different modules and the training strategy in details.

\subsection{Overall Pipeline}

Given a content image $x_c$ and a set of reference character images $y=\{y_1, y_2, \dots, y_K\}$ sharing the same font, our model aims to generate the character $y_c$ owning the same content with $x_c$ and the font style of the reference images. The overall framework is displayed in Fig.~\ref{fig_1}. It consists of a generator $G$ and a multi-task discriminator $D$.

The generator $G$ contains five parts as illustrated in Fig.~\ref{fig_1}. The content-encoder $E_c$ takes the content image as input to extract its structural representation $f_c$. It is pre-trained via VQ-VAE to obtain the component-wise latent codes $e_c$. The style encoder is designed to learn the style representations from reference images $y$. Specifically, it inputs each reference image independently to map them to the style latent vector $f_s=\{f_1, f_2, \dots, f_K\}$. A Local Style Aggregator (LSA) is followed afterward to transfer the styles onto the learned components. It works through a cross-attention module (CAM) with all the component-wise latent codes $e_c$ as queries and the representations of the reference characters $f_s$ as the keys and values. After the stylization of components, the content-related local styles $f_{sr-L}$ can be obtained by searching the most similar codes in $e_c=\{e_c^1, e_c^2, \dots, e_c^N\}$ with its representation $f_c$ spatially. The Global Style Aggregator (GSA) re-weights the style representations $f_s$ and sums up all of them channel-wisely to get the global representation $f_{sr-G}$. The weight is the normalized distance of the content representations between each reference character and the input character, as shown $f_r^c=\{f_1^c,f_2^c,\dots,f_K^c\}$ and $f_c$ in Fig.~\ref{fig_1}, respectively. Afterwards, a decoder is employed to generate target image $y_c$ with the concatenation features of the content features $f_c$, local style representation $f_{sr-L}$ and global style representation $f_{sr-G}$ as input.

During training, a multi-task discriminator is adapted to play the min-max game with the generator for distinguishing the real $\hat{y}_c$ from a fake $y_c$. To make sure the generated glyph has the correct style with reference samples and still retains the structure of the input content character, the discriminator outputs a binary classification of each character’s style and content category.

\subsection{Glyph Encoders and Component Decomposition}

\begin{figure}
    \centering
    \includegraphics[width=\linewidth]{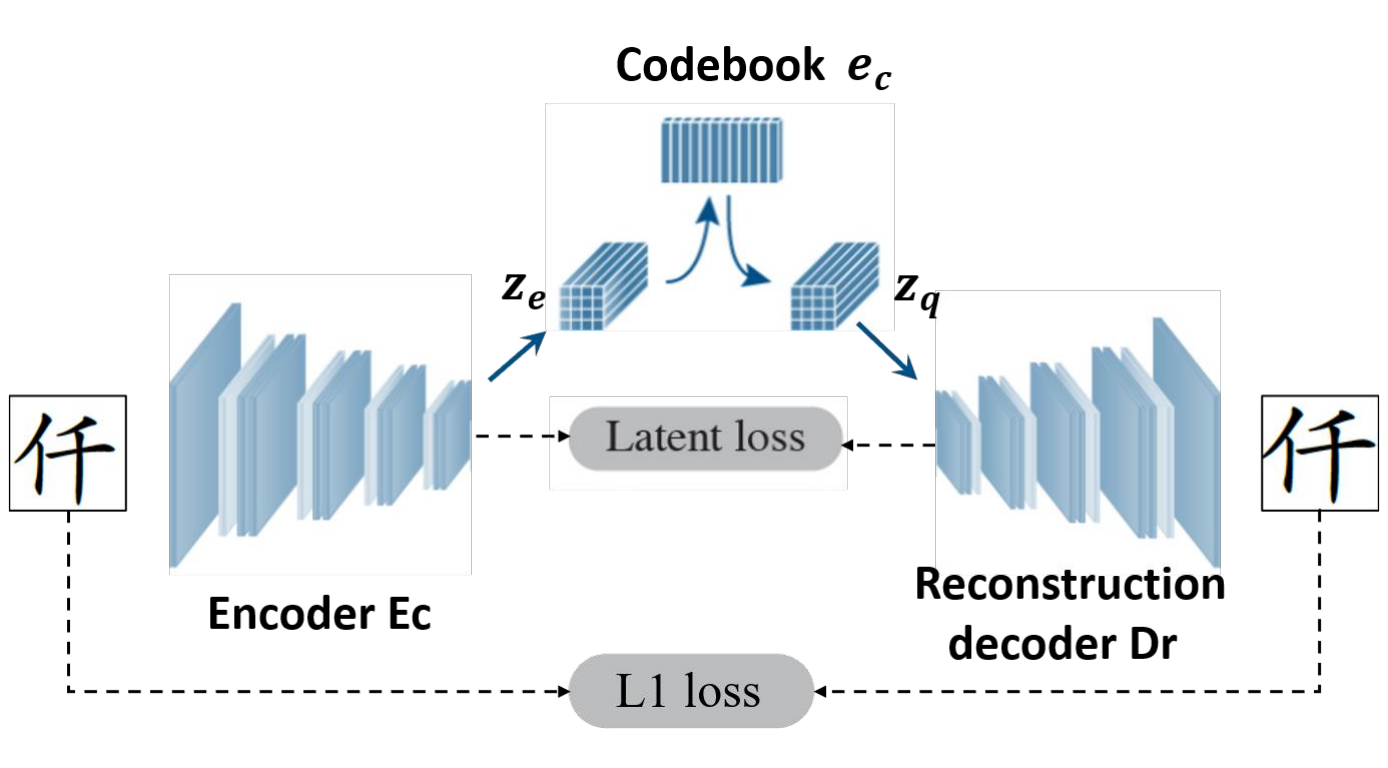}
    \caption{The glyph feature decomposing network for pre-training the content encoder and obtaining component representation.}
    \label{fig_2}
\end{figure}

The content encoder is pre-trained on the feature decomposing network and kept frozen for the font generation task. In the pre-training step, it is used to decompose the glyph images into component-wise latent codes, and trained on a certain set of character templates for reconstruction. The overview of the feature decomposing network is shown in Fig.~\ref{fig_2}. It contains a content encoder and a reconstruction decoder, connected by a discrete latent space using a latent codebook $e_c\in R^{N\times d}$. 

The encoder is built on CNNs and maps an input glyph image into latent
representations, $z_e \in R^{w\times h\times d}$, where d is the number of channels, and $h$ and $w$ represent the height and width of the feature maps, respectively. Subsequently, vector quantization~\cite{VQ-VAE} is used to discrete $z_e$ as follows:
\begin{equation}
    z_e^i=e_c^n, \quad {\rm s.t.} \quad n=\mathop{\arg\min}\limits_{n\in\{1,2,\dots,N\}}\Vert z_e^i-e_c^n\Vert_2^2,
    \label{eq1}
\end{equation}
where each spatial-wise elemental vector $z_e^i \in R^{d}$ in $z_e$ is replaced with the closest code vector by executing a nearest-neighbor lookup on codebook $e_c$, comprised of $d$-dimensional $n$ code vectors. Finally, the reconstruction decoder uses the matched codes $z_q \in R^{w\times h\times d}$ as an input and outputs a reconstructed glyph. 

To optimize this process, the encoder and codebooks are updated to minimize an objective $L_{pre}$ in Eq.~\eqref{eq2}, which refers to $L_1$-based reconstruction loss and latent loss $L_{lat}$ as follows:
\begin{equation}
    \begin{aligned}
    L_{pre} &= L_1+L_{lat}\\
    &=\Vert x_c-\hat{x}_c\Vert_1 +\alpha\Vert {\rm sg}[z_e]-e_c\Vert_2^2+ \beta\Vert z_e-{\rm sg}[e_c]\Vert_2^2,
    \end{aligned}
    \label{eq2}
\end{equation}
where $\rm{sg}$ is a stop-gradient operator that is defined as an identity at forward computation time and has zero partial derivatives. $\alpha$ and $\beta$ are the balancing hyperparameters. Experimentally, we set them as 1 and 0.1, respectively. The first term in Eq.~\eqref{eq2} makes sure no information gets lost in the forward pass. The second term updates the codebook variables by moving the embedding vectors $e_c$ towards the encoder outputs $z_e$. Meanwhile, the third term encourages the encoder output to match the targeted code vectors. There are no bypass connections between the encoder and decoder, so all the component-wise information of the glyph set processed by the encoder can be compressed in the latent space~\cite{VQ-VAE-2}. 

After this pre-training task, we fix the content encoder $E_c$ as well as the component codebook $e_c$ to build the font generation model. The style encoder $E_s$ has the same architecture as $E_c$, but it is trained from scratch.

\subsection{Local Style Aggregator}

Most recently proposed methods~\cite{Liu2022XMPFontSC,park2021few} extract local style representation of characters by gaining attention from input content and reference styles. It is inefficient, especially for creating a new font library that contains a large number of glyphs. Since each of the glyphs must go through the attention model for feature aggregation once. However, most of them actually share some local representation if they have the same strokes or radicals. Starting from this point, we design an LSA module, the core is a CAM block to transfer the styles onto all the component-wise codes $e_c$ instead of the input glyph.

The CAM block is built upon stacked multi-head transformers with component codes $e_c$ as queries and reference style vectors $f_r$ as keys and values. Formally, the feature maps $f_k\in R^{w\times h\times d}$ of $k$-th reference samples are reshaped along the channel axis, respectively, forming the sequential reference tokens $\tilde{f_r}\in R^{k\cdot w\cdot h\times d}$. For $i^{th}$ head, two linear projections $W_k^i\in R^{c\times d}$ and $W_v^i\in R^{c\times d}$ are applied on it to get the keys $K_i$ =$\tilde{f_r}W_k^{iT}$ and values $V_i$=$\tilde{f_r}W_v^{iT}$, respectively. Meanwhile, we acquire the query matrix $Q_i$ by linear projecting $e_c$ with a learning weight $W_e^i\in R^{c\times d}$.

The attention module operates on the queries $Q_i$, keys $K_i$, and values $V_i$ and generates weighted average vectors $\hat{V}_i$, which can be formulated as:
\begin{equation}
    \hat{V} = Attention(Q_i, K_i, V_i)={\rm softmax}\left(\frac{Q_iK_i^T}{\sqrt{c}}\right)V_i.
    \label{eq3}
\end{equation}

After obtaining the representations in each head, we concatenate all $\hat{V}_i$ along the channel dimension, and employ a linear projection to get the multi-head attention result $e_s$, as follows:
\begin{equation}
    e_s = Multi-head(Q, K, V)={\rm concat}\left(\hat{V}_1, \hat{V}_2,\dot,\hat{V}_m\right)W_o,
    \label{eq4}
\end{equation}
where $m$ is the number of total attention head, and $W_o$ is the learning weight.

After we get the stylized component-wise codes $e_s=\{e_s^1, e_s^2, \dots, e_s^N\}$. For each input glyph, its local representation $f_{sr-L}$ can be aggregated from $e_s$ by searching for the closest label $n$ in $e_c$ for each spatial element in $f_c$ as expressed in Eq.~\eqref{eq1}, and then selecting the corresponding style representation $e_s^n$ for replacement.  

\subsection{Global Style Aggregator}

The original style representation $f_k$ for each reference sample is able to provide more global information for the font generation, which can control the size, the stroke space, etc. So, we aggregate their features directly as the global representation.

As illustrated in~\cite{9098082}, the style features and content features of references are not entirely independent. From a human’s perceptual observation, there are some correlations that exist between different characters. If the input glyph has exactly the same radical or structure with some references, their style representations should be highly referred to during the transfer process. Therefore, we re-weight the individual style features $f_k$ of each reference glyph based on the content feature similarity. We extract the content features $f_r^c=\{f_1^c, f_2^c,\dots, f_K^c\}\in R^{w\times h\times d}$ of each reference and reshape them to $\tilde{f_k^c} \in R^{w\cdot h\times d}$, meanwhile, the content features$f_c\in R^{w\times h\times d}$ of content image are reshaped into $\tilde{f_c}\in R^{w\cdot h\times d}$. The glyph similarity is determined by the normalized cross-correlation measurement as:
\begin{equation}
   W_{kj} =\frac{\tilde{f}_k^c\cdot \tilde{f}_c}{\left\| \tilde{f}_k^c\right\|*\left\| \tilde{f}_c\right\|}, j\in\{1,2,\dots, d\}.
   \label{eq5}
\end{equation}
$W_{kj}$ is a scalar that represents the similarity between $k_{th}$ reference and the input glyph on $j_{th}$ channel. Then, we normalize it by considering $k$ samples on each channel as:
\begin{equation}
   \overline{W}_{kj} ={\rm softmax}\left(\frac{exp(aW_{kj})}{\sum_{k=1}^Kexp(W_{kj})}\right).
   \label{eq6}
\end{equation}

This weight is then applied to the reference style representations $f_r$, where $j_th$ channel feature $f_k^j$ is weighted by $\overline{W}_{kj}$, to aggregate the global representation $f_{sr-G}$ as expressed in:
\begin{equation}
   f_{sr-G}={\rm concat}_j\left(\sum_{k=1}^K\overline{W}_{kj}f_k^j\right).
   \label{eq7}
\end{equation}
This global style representation $f_{sr-G}$ is concatenated with the content features $f_c$ and local style representation $f_{sr-L}$, and then input to the decoder to generate the target font.

\subsection{Training}

We train our model to generate the image $y_c$ from an input content image $x_c$ and a set of reference glyph images $y=\{y_1, y_2, \dots, y_K\}$. The content encoder is pre-trained and kept fixed during this period. The losses consist of three parts, an adversarial loss, a matching loss, and a style contrast loss. The adversarial loss and matching loss are used between the generated results $y_c$ and the ground truth $\hat{y}_c$. While the style contrast loss is designed to learn distinguished styles for components.

\textbf{Adversarial loss.}
To make the model generate plausible images, we employ a multi-head projection discriminator $D_{s,c}$ for style label $s$ and character label $c$ in our framework. The loss function is implemented as the hinge GAN loss~\cite{2018Self}:
\begin{equation}
\begin{aligned}
\mathcal{L}_{adv}^D = &-\mathbb{E}_{\hat{y_c}\sim p_{data}}min(0,-1+D_{s,c}(\hat{y_c}))\\
  &- \mathbb{E}_{y_c\sim p_{G}}min(0,-1-D_{s,c}(y_c))\\
\mathcal{L}_{adv}^G = & - \mathbb{E}_{y_c\sim p_{G}}D_{s,c}(y_c),
\end{aligned}
\label{eq8}
\end{equation}
where $p_G$ denotes the set of generated images, and $p_{data}$ denotes the set of real glyph images.

\textbf{Matching loss.}
To make the generated character $y_c$ learn pixel-level and feature-level consistency with ground truth $\hat{y_c}$, we employ an $L_1$ loss on both image and image features as follows:
\begin{equation}
\begin{aligned}
&\mathcal{L} _{img} = \mathbb{E}[\left \| y_c-\hat{y_c}  \right \|_{1}]\\
&\mathcal{L} _{feat} = \mathbb{E}[\sum_{l=1}^{L}\left \| f^{(l)}(y_c) - f^{(l)}(\hat{y_c})\right \| _{1}],
\end{aligned}
\label{eq9}
\end{equation}
where $f_l$($y_c$) and $f_l$($\hat{y_c}$) represent the intermediate features in the $l_th$ layer of $D$.

\textbf{Style contrast loss.} 
Given two different reference sets but sharing the same font style, the associated stylized components for these sets should be the same. However, if the reference character sets have different font styles, the component styles should be distinguished. So, we formulate our style contrastive loss as follows:
\begin{equation}
\begin{split}
  \mathcal{L}_{cst}&= \\
  -log&\left(\frac{exp\left(\sum\limits_{n=1}^Ne_s^{nT}e_{s+}^{n}\right)}{exp\left(\sum\limits_{n=1}^Ne_s^{nT}e_{s+}^{n}\right)+\sum\limits_{N_s^-} exp{\left(\sum_{n=1}^Ne_s^{nT}e_{s-}^{n}\right)}}\right),
   \label{eq10}
\end{split}
\end{equation}
In detail, for each certain font style $s$ and is corresponding codebook $e_{s}$, where $e_{s+}$ is the positive codebook pair of $e_{s}$ whose references have the same font style but different content, and $e_{s-}$ is the negative codebook pair share different font style. $N_s^-$ is the number of negative style samples, for $n$-th codebook entries $e_s^n$ in codebook $e_s\in R^{N\times d}$, We can always find the positive pair $e_{s+}^n$ and negative pair set $\{e_{s_1-}^n,e_{s_2-}^n,\dots,e_{s_{neg}-}^n\}$ in the corresponding position. 

\textbf{Overall objective loss.}
Finally, we optimize the whole model by the following full objective function:
\begin{equation}
\min\limits_{E_s,G}\max\limits_{D}\mathcal{L}_{adv}^D +\mathcal{L}_{adv}^G + \lambda_1\mathcal{L}_{img} \\
+\lambda_2\mathcal{L}_{feat}+\lambda_3\mathcal{L}_{cst}.
\label{eq11}
\end{equation}
The $\lambda_1$, $\lambda_2$ and $\lambda_3$ is weighting hyperparameter. We set them to 1, 1, and 0.1, respectively.

\section{Experimental Results}

\subsection{Experiment Setup}

\textbf{Dataset.} We collected 386 Chinese fonts and generated 3,500 Chinese characters for each font according to the first-level Chinese Character Table. All character pictures are normalized to 128$\times$128. A font template is randomly selected from the 386 fonts for extracting the glyph content features and kept fixed throughout the training and test~\cite{MC-GAN}. This font set is also used for pre-training the VQ-VAE to get a set of common parts codebooks.

The Chinese training set contains 370 fonts and 3,000 Chinese characters, denoted as the seen font seen character (SFSC) set. To verify our method, we evaluate the font generation ability on two test sets: One is the rest of the 15 unseen fonts with 3,000 seen characters per font, denoted as the seen font unseen character (UFSC) set; the other is the remaining 15 unseen fonts with 500 unseen characters per font, denoted as the unseen font unseen character (UFUC) set.

\textbf{Implementation details.}
The main training process of our method is divided into two parts. First, a component-wise codebook is trained with 3,000 Chinese characters sharing the same font. We set the embedding dimension $d$ to 256, the batch size to 256, and the iteration steps to 50,000. Then, for training the whole font generation model, we set the batch size to 48, the number of attention heads to 8 in three stacked transformer layers, and iteration steps to 500,000.

\textbf{Evaluation metrics.} To evaluate the font generation quality, we use the following five metrics~\cite{Look}, root-mean-square deviation (RMSE), structural similarity (SSIM), learned perceptual image patch similarity (LPIPS)~\cite{LPIP}, Fr\'{e}chet inception distance (FID)~\cite{FID}, and User Study. The User Study is conducted by 10 volunteers, who observe the reference samples and vote for the best generation result from all the comparison methods.

\subsection{Ablation Studies}
We perform three ablation studies to evaluate the effectiveness
of our proposed models on the few-shot Chinese font generation task. These experiments are conducted on the UFUC dataset.

\subsubsection{The effect of component codebook size}
The component codebook size reflects the complexity of the script. We set it as different values for training the content encoder and acquiring component-wise representation in VQ-VAE. Then, they are fixed for font generation. The results are displayed in Tab.~\ref{tab1}. We can see that when the codebook size exceeds 100, the performance on each metric has limited improvement. That illustrates using 100 components is sufficient to depict the Chinese characters. When we train the model on English script, we find that by only using 15 codes, the model can generate very good results. Therefore, the size of the component codebook has a relationship with the scripts. More complex scripts require larger codebook sizes. In the following experiments, We set the Chinese component codebook size to 100.

\begin{table}\tiny
\caption{Ablation study of component codebooks.}
\label{tab1}\centering
\resizebox{\linewidth}{!}{
\begin{tabular}{ c c c c c }
\hline
{Codebook size} & {SSIM$\uparrow$} & {RMSE$\downarrow$} & {LPIPS$\downarrow$} & {FID$\downarrow$} 
 \\ 
\hline
 50  & 0.512 & 0.422 & 0.343 &144.2 \\
\hline
 80  & 0.546 & 0.407 & 0.296 &120.7\\
\hline
 100  & 0.566 & 0.390 & 0.282 &110.1\\
 \hline
 120  & 0.568 & 0.392 & 0.278 &109.5\\
 \hline
 150  & 0.567 & 0.388 & 0.282 &112.4 \\
\hline
\end{tabular}}
\end{table}

\begin{figure}
    \centering
    \includegraphics[width=0.8\linewidth]{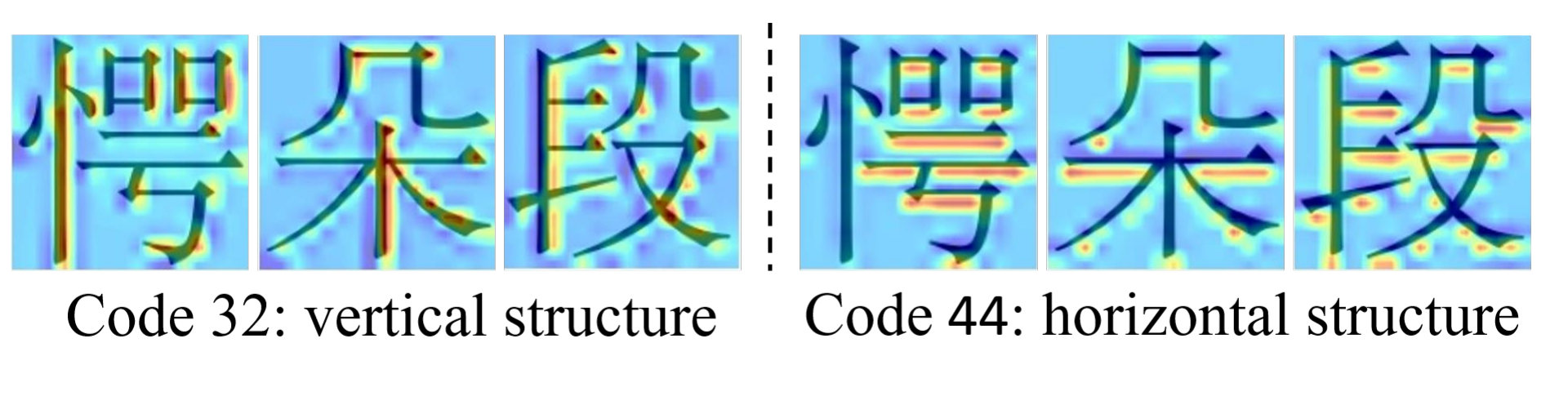}
    \caption{The visualization of latent codes.}
    \label{fig_8}
\end{figure}

We select a certain code from the codebook to compute its similarity with features of different glyphs spatially. The visualization results of the 32$_{th}$ code and 44$_{th}$ code are present in Fig.~\ref{fig_8}. We can see the 32$_{th}$ code mainly extracts vertical components and 44$_{th}$ extracts horizontal components. It demonstrates the VQ-VAE strategy can generate ``components'' automatically.

\begin{figure}
    \centering
    \includegraphics[width=0.9\linewidth]{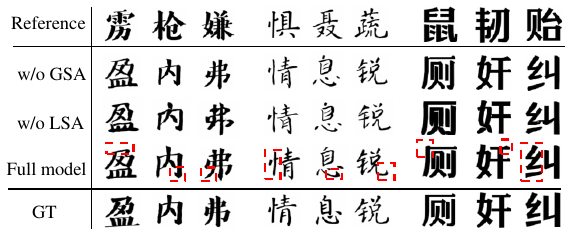}
    \caption{Qualitative font generation results with and without LSA and GSA modules. The first and last rows are the three references and the ground truth per font, respectively. The middle rows are generated results correspondingly. The red dashed boxes point out details that were better generated by our full proposed model.}
    \label{fig_3}
\end{figure}

\subsubsection{The effect of each module}

To illustrate the influence of the GSA and LSA modules of our method, we conduct the ablation study with and without it in the model. By removing the LSA, the component-wise representation is not required. The model degrades to a global content-style disentanglement architecture, with the aggregated style representation only conditioned on content similarity. When removing the GSA part, the model focuses more on the details of the stroke but loses the global style control. 

\begin{table*}\tiny
\caption{Quantitative comparison results on UFSC and UFUC datasets}
\label{tab3}\centering
\resizebox{0.75\linewidth}{!}{
\begin{tabular}{ c|c| c c c c c}
\hline
Dataset& Method &SSIM$\uparrow$ & RMSE$\downarrow$ &LPIPS$\downarrow$& FID$\downarrow$ & User Study$\uparrow$\\
\hline
\multirow{7}*{UFSC}
 & FUNIT~\cite{FUNIT} & 0.512 & 0.426  & 0.361  & 161.8  &1.2\% \\
 & LF-FONT~\cite{park2021few}&0.548 & 0.406 & 0.334 &123.9 & 3.9\% \\
 & AGIS-NET~\cite{AGIS} &0.529 & 0.409 & 0.302 & 145.0& 10.3\% \\
 & DG-FONT~\cite{DG-FONT} & 0.517  &0.419 & 0.312 & 149.0& 11.6\%\\
 & MX-FONT~\cite{park2021multiple} &0.523 & 0.407 & 0.297 & 94.5 & 14.4\%\\
 & FS-FONT~\cite{Tang2022FewShotFG} & 0.570& 0.379 & 0.327 & 172.6 & 21.5\%\\
 & Ours &\textbf{0.636}& \textbf{0.341} & \textbf{0.225}&\textbf{93.7} & \textbf{58.6\% } \\
\hline
\multirow{7}*{UFUC} 
& FUNIT~\cite{FUNIT} &0.468& 0.472 & 0.379 & 153.4& 3.4\%  \\
 & LF-FONT~\cite{park2021few} & 0.481 & 0.462& 0.377&136.7 &4.8\%\\
 & AGIS-NET~\cite{AGIS} &0.550& 0.400 & 0.310 & 137.2 &  11.5\%\\
 & DG-FONT~\cite{DG-FONT}& 0.493 & 0.397 & 0.312 & 136.7 &  10.8\%\\
 & MX-FONT~\cite{park2021multiple} &0.478& 0.447 & 0.333& 114.8 & 15.2\%\\
  & FS-FONT~\cite{Tang2022FewShotFG} &0.418 & 0.474 & 0.377& 173.7 & 19.5\%\\
 & Ours &\textbf{0.566} & \textbf{0.390} & \textbf{0.282} &\textbf{110.1} & \textbf{54.3\%}  \\
\hline
\end{tabular}}
\end{table*}

We evaluate the two models and the proposed full model on our collected UFUC dataset. Fig.~\ref{fig_3} displays some comparison results. The results with and without these modules have obvious differences. The model with LSA could better capture the details of reference strokes, while the sizes of generated characters and the inner spaces are not always consistent with the given references. The model with only GSA has the opposite functions. Most strokes are not generated well. The full model combines LSA and GSA, which achieves better performance (see the details in red dash boxes of Fig.~\ref{fig_3}). Quantitative results in Tab.~\ref{tab2} further confirm the importance of the two modules. The LSA model assists more than GSA, which demonstrates that the style of glyphs mainly lies in the local details. From these results, we conclude that LSA and GSA are both essential in our model and they enable the model to capture diverse local styles and also control the global styles. 

\begin{table}
\caption{Ablation study of different models on UFUC dataset.}
\label{tab2}\centering
\resizebox{\linewidth}{!}{
\begin{tabular}{ c c c c c c}
\hline
{Methods} & {SSIM$\uparrow$} & {RMSE$\downarrow$} & {LPIPS$\downarrow$} & {FID$\downarrow$} & User Study$\uparrow$ 
 \\ 
\hline
 w/o LSA & 0.496 & 0.422 & 0.318 &145.0 & 7.3\%  \\
\hline
 w/o GSA & 0.532 & 0.402 & 0.293 &132.1 & 20.3\% \\
\hline
Full model  & \textbf{0.566} & \textbf{0.390} & \textbf{0.282} &\textbf{110.1} & \textbf{72.4\%} \\
\hline
\end{tabular}}
\end{table}

\subsubsection{The effect of reference numbers}

\begin{figure}
    \centering
    \includegraphics[width=0.8\linewidth]{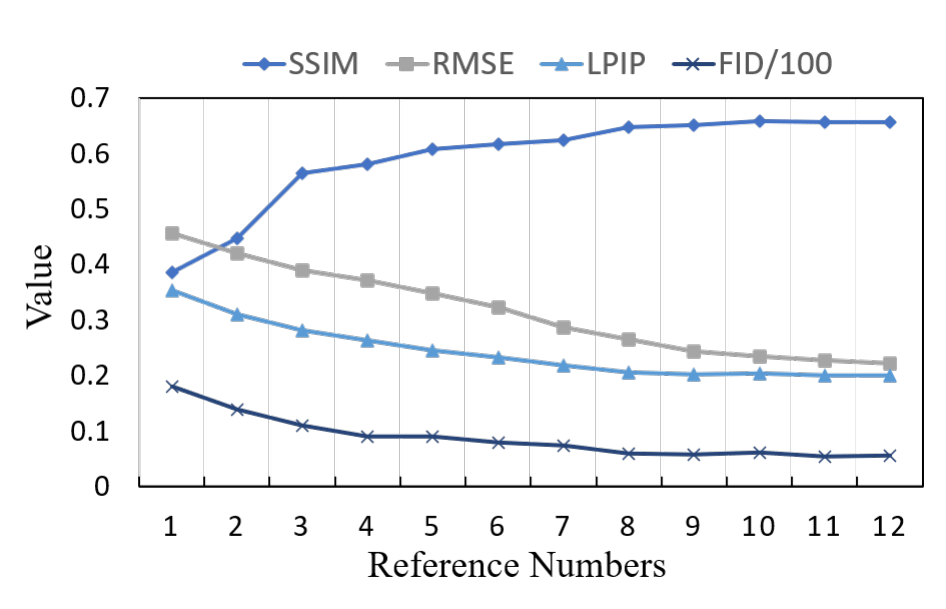}
    \caption{Performance changes of our model by varying numbers of the reference samples.}
    \label{fig_4}
\end{figure}

Intuitively, given more reference samples, the generated font better resembles a realistic one. Fig.~\ref{fig_4} shows the effect of reference numbers. We can see that when the number of references increases from 1 to 8, the generated characters have an incremental performance improvement. The figure shows that it becomes stable when given more references. However, there are diminishing returns after the first few references. More details can be found in Appendix.

\begin{figure*}
    \centering
    \includegraphics[width=\linewidth]{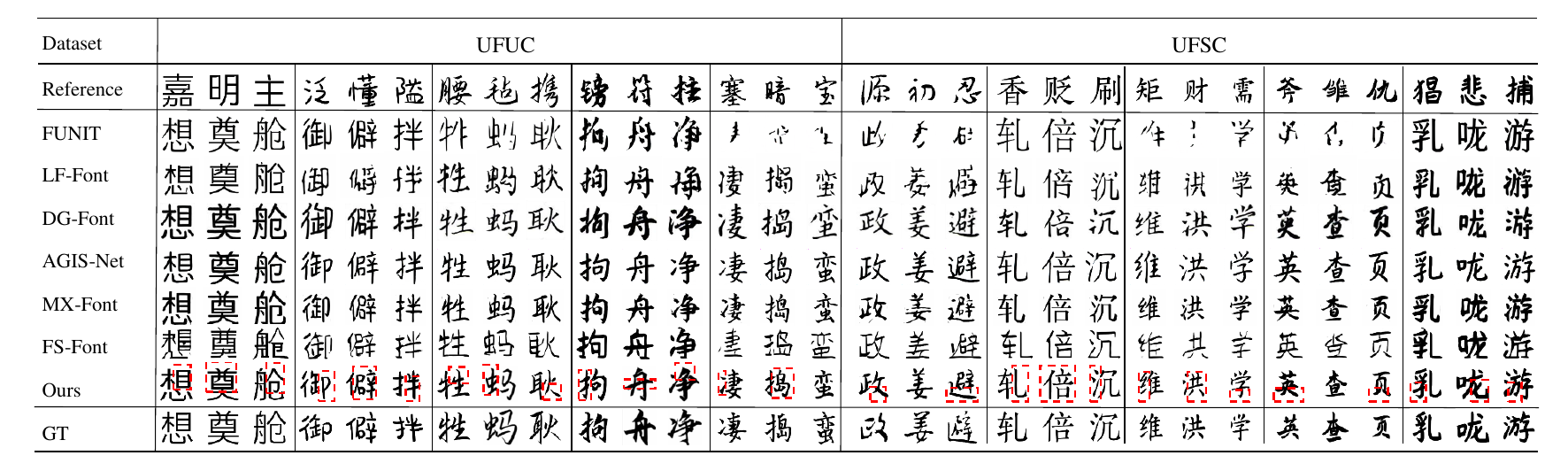}
    \caption{FFG results of each method on UFUC and UFSC dataset. We represent the generated samples of five different kinds of fonts, given three references and three content images per font. The red boxes represent the better generation details.}
    \label{fig_5}
\end{figure*}

\subsection{Comparing with SOTA Methods}
We compare our method with five SOTA AFFG methods. 1) FUNIT~\cite{FUNIT} used two different encoders with the AdaIN module to generate a new image with mixed content and style. It belongs to the global style-content disentanglement FFG method. 2) MX-Font~\cite{park2021multiple} extracts multiple style features not explicitly conditioned on component labels, but is automatically by
designed multiple experts to represent different local concepts, e.g., left-side sub-glyph. It shows good generalization ability to unseen languages. 3) LF-Font~\cite{park2021few} proposed localized style representation inspired by low-rank matrix factorization, which also extracts component-wise features. 4) DG-Font~\cite{DG-FONT} is an unsupervised network that used a Deformable Convolution in the Generator to deform and transform the character of one font to another. It achieved a good effect on cursive characters. 5) AGIS-net~\cite{AGIS} used two different decoders to generate images with character shapes and then the texture information successively. It can transfer the font as well as the text effects. 6) FS-Font~\cite{Tang2022FewShotFG} requires fixed content-reference mapping to transfer Fine-Grained local styles. We retrain all the models by our dataset using their original training hyper-parameters and strategies. The reference samples are set to 3 in the training and test phases for all the methods. The qualitative and quantitative comparison results are shown in Fig.~\ref{fig_5} and Tab.~\ref{tab3}, respectively.

Tab.~\ref{tab3} shows that our method outperforms previous SOTA methods with significant gaps in all the metrics. We can observe that FUNIT fails to capture diverse styles. Only when the content and reference styles are visually close, it can generate a structured character. That illustrates the universal style representation strategy is not a good choice for AFFG. LF-Font generates fonts with worse visual quality since the reference numbers are very limited. Its performance is unstable as they are prone to loss of strokes or distortion of components for certain styles. DG-Font may fail to perform style transfer especially when the reference style significantly differs from that of the content. Additionally, it generates glyphs containing characteristic artifacts. AGIS-Net is more stable to generate the font of different styles and complement structures. It could better transfer the global features such as the size and the thickness of strokes. MX-Font generates better stylized characters than other methods, but it has a higher probability of failing to generate fine-grained style features compared with our method. The generation results of FS-Font are also unsatisfactory, especially on unseen references, due to the randomly selected content-reference pairs.

\subsection{Applying to Other Languages}

\begin{figure}
    \centering
    \includegraphics[width=\linewidth]{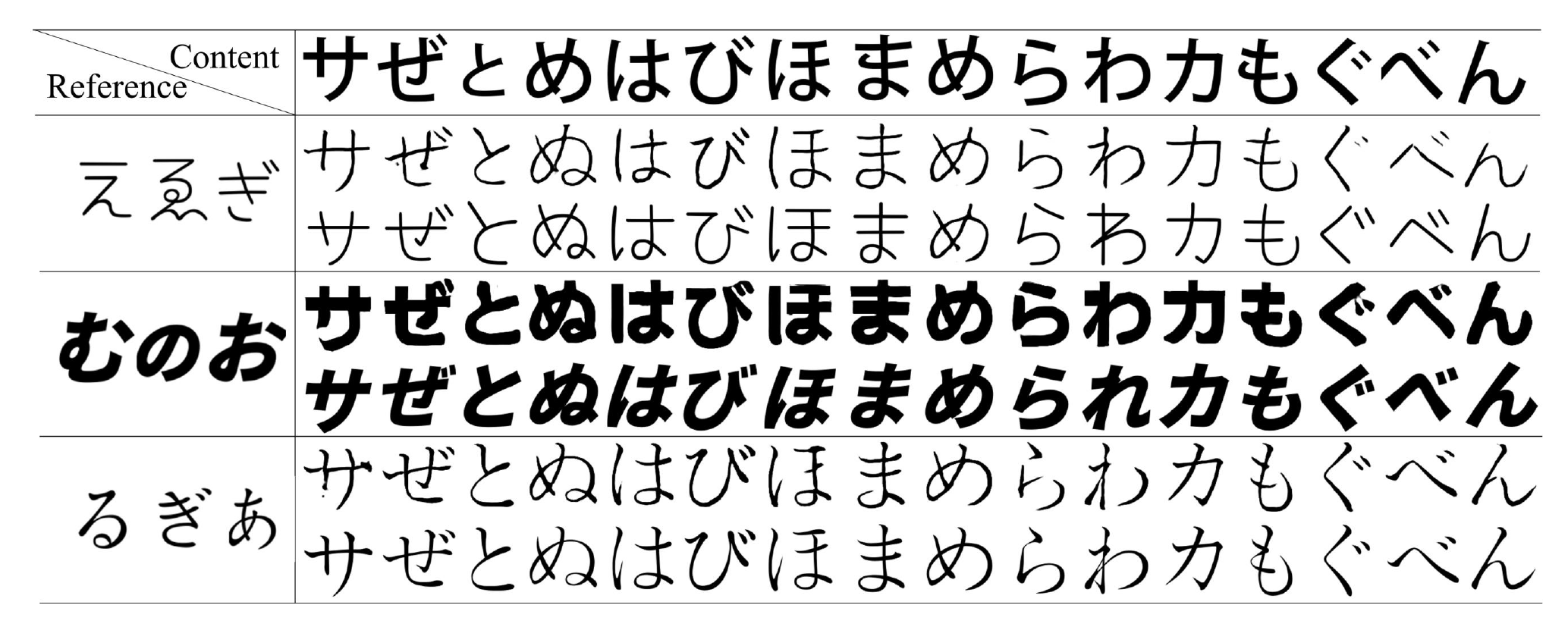}
    \caption{FFG results of Japanese scripts. For each style, the upper and lower rows respectively represent the model generation results and GT. }
    \label{fig_12}
\end{figure}
\begin{figure}
    \centering
    \includegraphics[width=\linewidth]{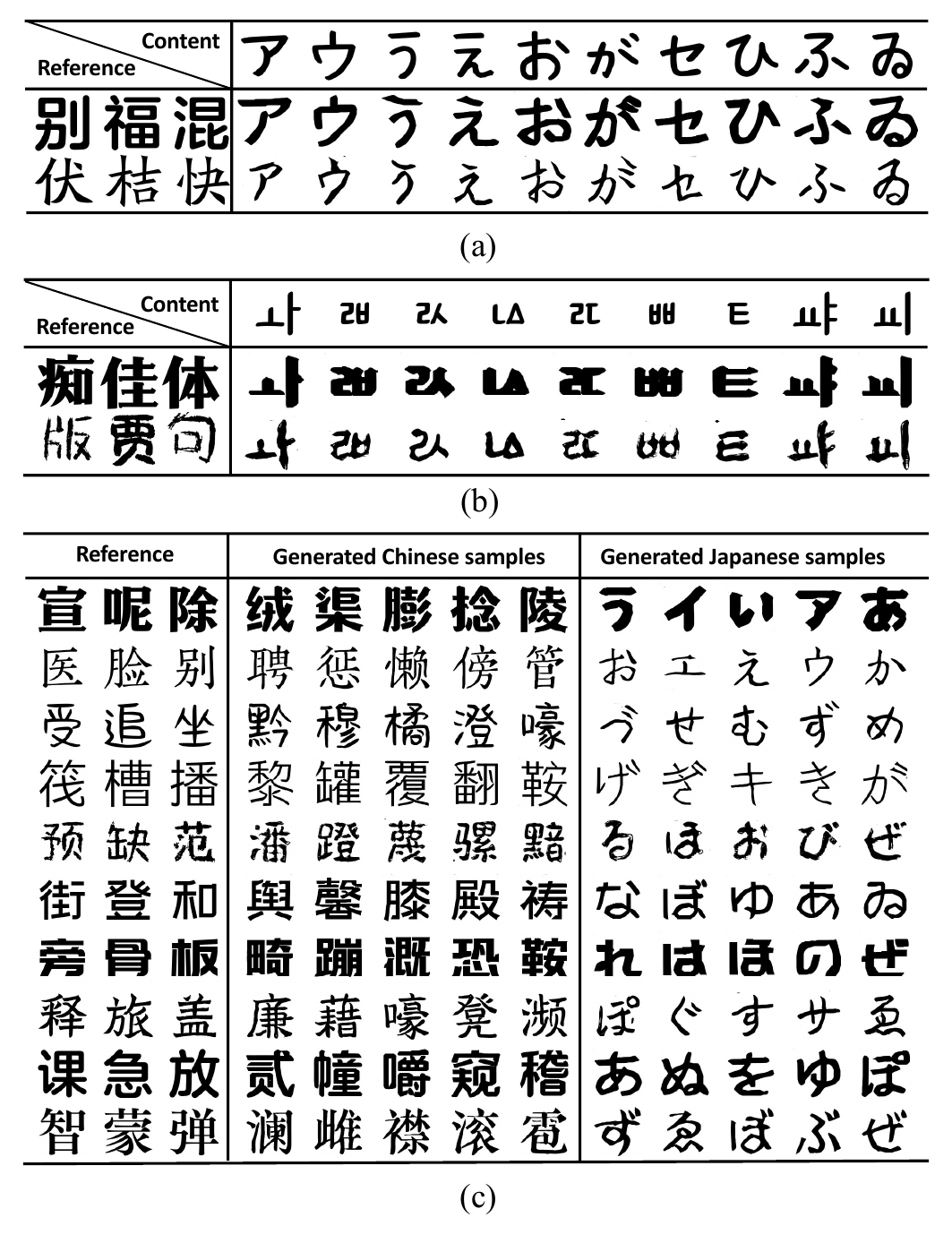}
    \caption{Cross-language inference on (a) Japanese glyphs and (b) Korean glyphs. (c) Chinese results generated under few-shot and Japanese generated results under zero-shot manner.}
    \label{fig_11}
\end{figure}

To verify the generalization of our method for other languages, we collected 60 fonts of 125 Japanese characters. 50 fonts with 100 glyphs per font are randomly selected for training, and the rest for test. Some results are displayed in Fig.~\ref{fig_12}. Since the character structures of the Japanese are much simpler than Chinese, the generated results approach the ground truth more. Additionally, our method can transfer styles onto unseen content, including cross-linguistic glyphs as shown in Fig.~\ref{fig_11}.


\section{Conclusion}

We present a novel AFFG approach in this paper, which could aggregate global and local styles of limited references. The designed LSA and GSA modules were assigned different tasks to capture the local details and global structures of the font, respectively. The LSA adopted a cross-attention mechanism for transferring styles onto all of the self-learned components without manual definition. While the GSA utilized the similarity of different glyphs, and then used it to guide the global styles combination of all references. Experiments demonstrated the effectiveness of both modules, and also showed our proposed method significantly outperformed other methods quantitatively and qualitatively. However, our method is limited to two aspects. If given only one reference or very fancy font, e.g., decoration, or shadow, the generated results are unsatisfactory.

\section*{Acknowledgments}
This work is supported by the Open Project Program of the National Laboratory of Pattern Recognition(NLPR) (No.202200049). We thank all anonymous reviewers for their constructive comments on this paper.

\section*{Appendix}
Besides the results and discussions presented in the original paper, we provide more experiments and results in this appendix, which further explain the effectiveness and generality of our method.

\subsection*{The effect of different reference characters on the generated results}

\begin{figure}
    \centering
    \includegraphics[width=\linewidth]{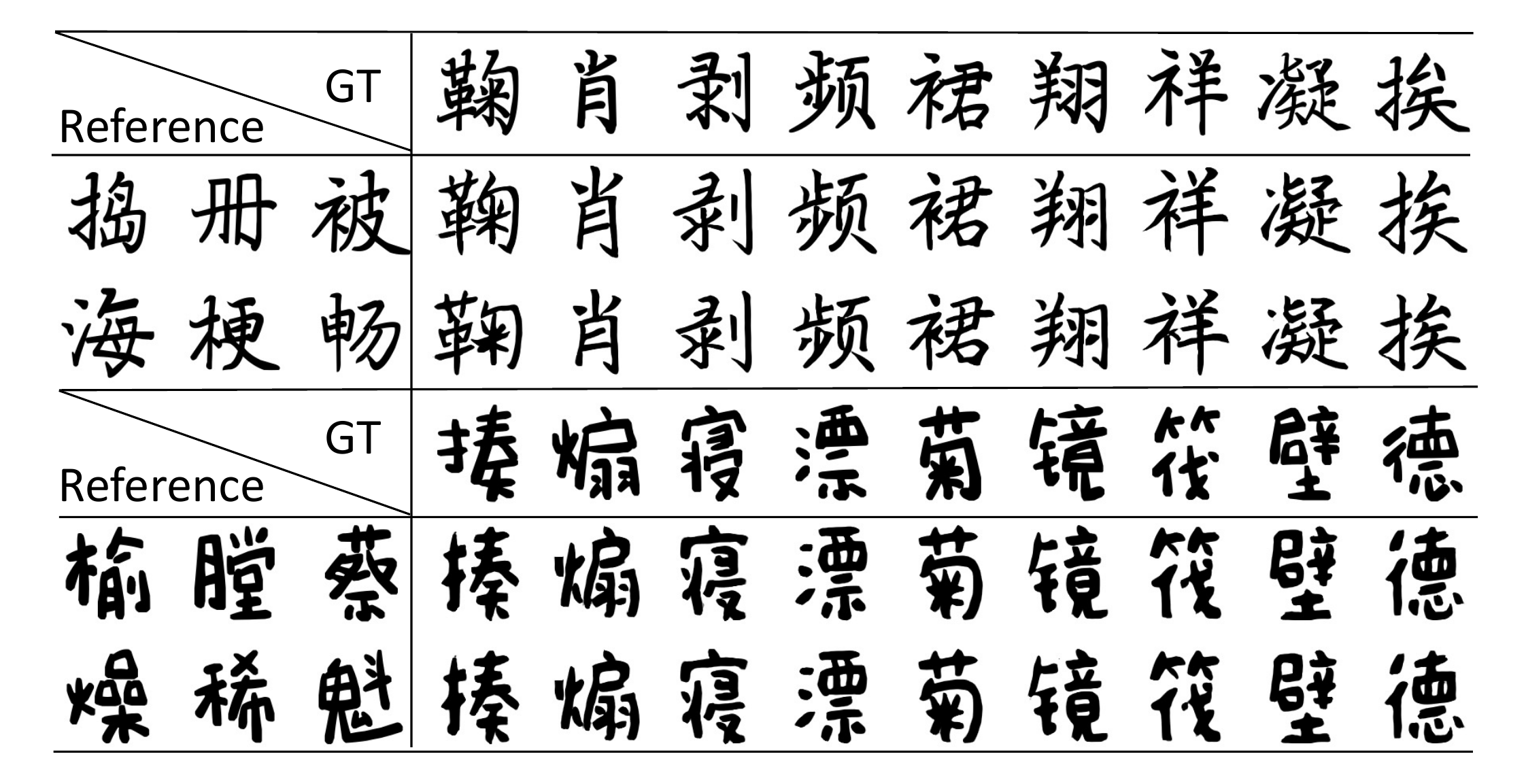}
    \caption{Generating the same target character, referring to characters with different contents from the same style font.}
    \label{fig_appendix_1}
\end{figure}

The font style is an aesthetic concept and has nothing to do with specific characters. In the process of few shot font generation, given reference characters with different content, the provided style representation should be the same. In our training process, we randomly select $k$ style reference characters for generating characters each time. During the inference phase, we select characters with different content in the same style as the reference set. The generated results are shown in Fig.~\ref{fig_appendix_1}. There is a subtle difference between the overall style and the local style of the generated results.

\subsection*{The qualitative results using different numbers of references} 
\begin{figure}
    \centering
    \includegraphics[width=\linewidth]{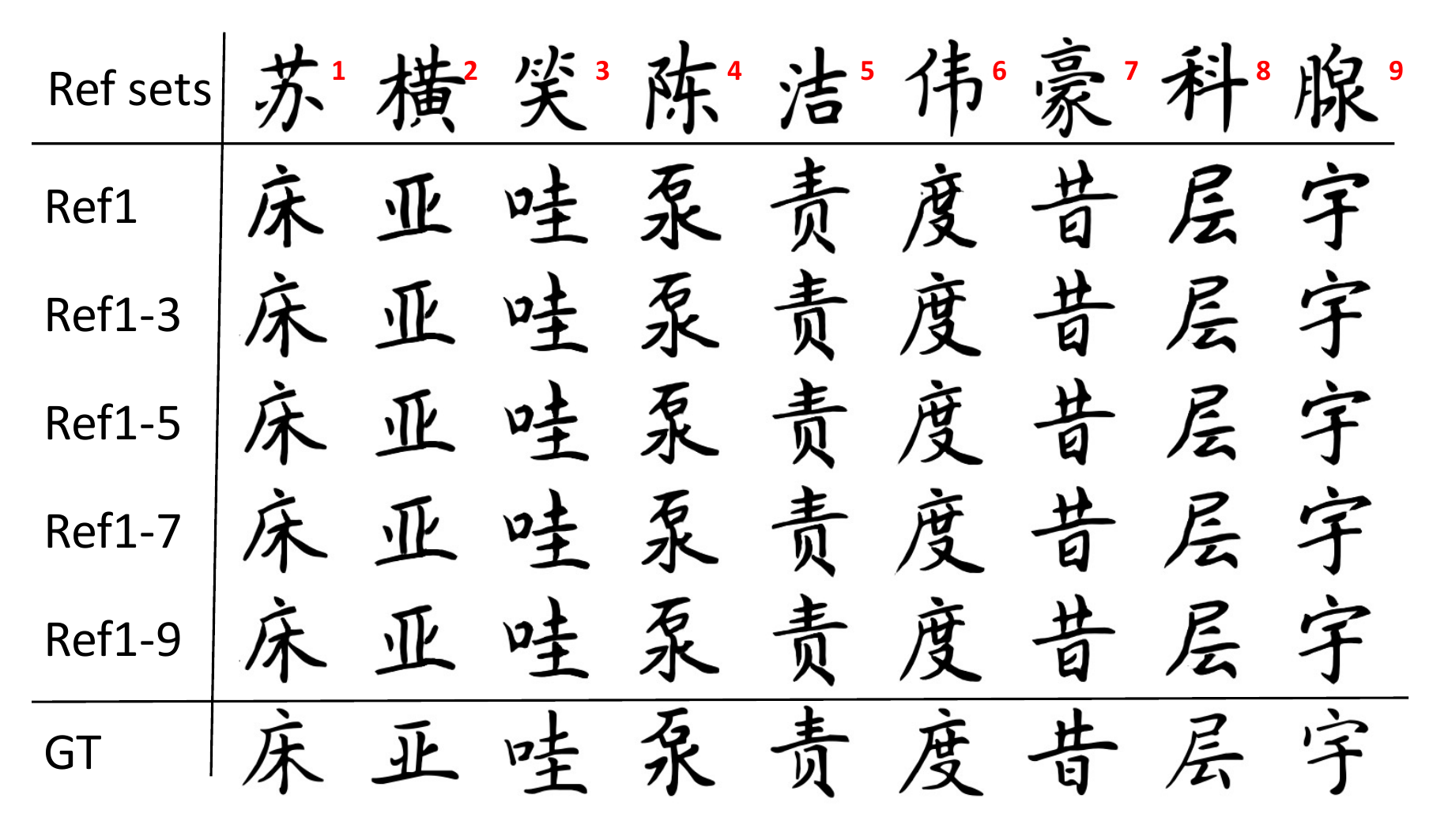}
    \caption{Generated samples by varying reference set size. Each middle row shows the samples generated by given different number (1, 3, 5, 7, 9) of references. The number (1 to 9) represents the order of each character to be used. The target glyphs are displayed in the bottom row.}
    \label{fig_appendix_2}
\end{figure}

Intuitively, the larger the number of given reference characters, the richer and more accurate the extracted style features should be. In the inference stage, we change the number of reference characters and observe the impact on the generated results. It can be observed from Fig.~\ref{fig_appendix_2} that the overall and local quality of generated characters is roughly positively correlated with the number of reference characters. The quality improvement is more obvious when the number increases from 1 to 5, and the improvement beyond this number is limited. 
In our paper, the number of reference images is set to 3, and the model training and qualitative and quantitative comparisons with other methods are completed with this setting.

{\small
\bibliographystyle{ieee_fullname}
\bibliography{egbib}
}

\end{document}